\documentclass{article} 

\usepackage{iclr2026_conference,times}

\usepackage{amsmath,amsfonts,bm}

\def\eqref#1{equation~\ref{#1}}

\def\1{\bm{1}}

\DeclareMathAlphabet{\mathsfit}{\encodingdefault}{\sfdefault}{m}{sl}
\SetMathAlphabet{\mathsfit}{bold}{\encodingdefault}{\sfdefault}{bx}{n}

\usepackage{comment}
\usepackage{hyperref}

\usepackage[utf8]{inputenc}

\usepackage{enumitem}
\setlist[enumerate,itemize]{leftmargin=*, labelindent=0pt}
    \usepackage{siunitx}
\usepackage{graphicx}
\usepackage{wrapfig}
\usepackage{booktabs}
\usepackage{tabularx} 
\usepackage{url}      
\usepackage[table]{xcolor}  

\usepackage[T1]{fontenc}     
\usepackage{lmodern}         
\usepackage[english]{babel} 

\definecolor{lightgreen}{RGB}{0,150,0}  
\definecolor{lightred}{RGB}{204,0,0}    

\definecolor{red10}{RGB}{252,174,145}   
\definecolor{red20}{RGB}{251,106,74}    

\newcommand{\lantern}{\textsc{LanteRn}}

\definecolor{tagorange}{HTML}{D35400} 
\definecolor{tagpurple}{HTML}{6C3483} 
\definecolor{taggreen}{HTML}{27AE60}  
\definecolor{tagblue}{HTML}{3498DB}

\title{\lantern:\\Latent Visual Structured Reasoning}

\author{%
$ $ \\
  \makebox[\textwidth][c]{%
    \begin{tabular}{c}
      \textbf{André G. Viveiros}\thanks{Corresponding author: \texttt{andre.viveiros@tecnico.ulisboa.pt}.} $^{1,2}$ \quad\quad \textbf{Nuno Gonçalves}$^{1,2,3}$ \\[0.5ex]
      \textbf{Matthias Lindemann}$^{1}$ \quad\quad \textbf{André F. T. Martins}$^{1,2}$ \\[2ex]
      \textnormal{$^1$ Instituto de Telecomunicações} \\
      \textnormal{$^2$ Instituto Superior Técnico, Universidade de Lisboa} \\
      \textnormal{$^3$ Carnegie Mellon University}
    \end{tabular}
  }
}

\iclrfinalcopy 
\begin{document}

\maketitle

\begin{abstract}
While language reasoning models excel in many tasks, \textit{visual} reasoning remains  challenging for current large multimodal models (LMMs). As a result, most LMMs default to verbalizing perceptual content into text, a strong limitation for tasks requiring fine-grained spatial and visual understanding. 
While recent approaches take steps toward \emph{thinking with images} by invoking tools or generating intermediate images, they either rely on external modules, or incur unnecessary computation by reasoning directly in pixel space.\\
In this paper, we introduce \lantern{}, a framework that enables LMMs to interleave language with compact latent visual representations, allowing visual reasoning to occur directly in latent space. \lantern{} augments a vision-language transformer with the ability to generate and attend to continuous visual ``thought'' embeddings during inference. We train the model in two stages: supervised fine-tuning to ground visual features in latent states, followed by reinforcement learning to align latent reasoning with task-level utility. 
We evaluate \lantern{}{} on three perception-centric benchmarks (VisCoT, $V^\star$, and Blink), observing consistent improvements in visual grounding and fine-grained reasoning. These results suggest that internal latent representations provide a promising direction for more efficient multimodal reasoning. The code is available at \href{https://github.com/GuilhermeViveiros/LanteRn}{LanteRn}.

\end{abstract}

\vspace{-0.3cm}
\begin{figure}[h]
\centering
  \includegraphics[width=0.72\linewidth]{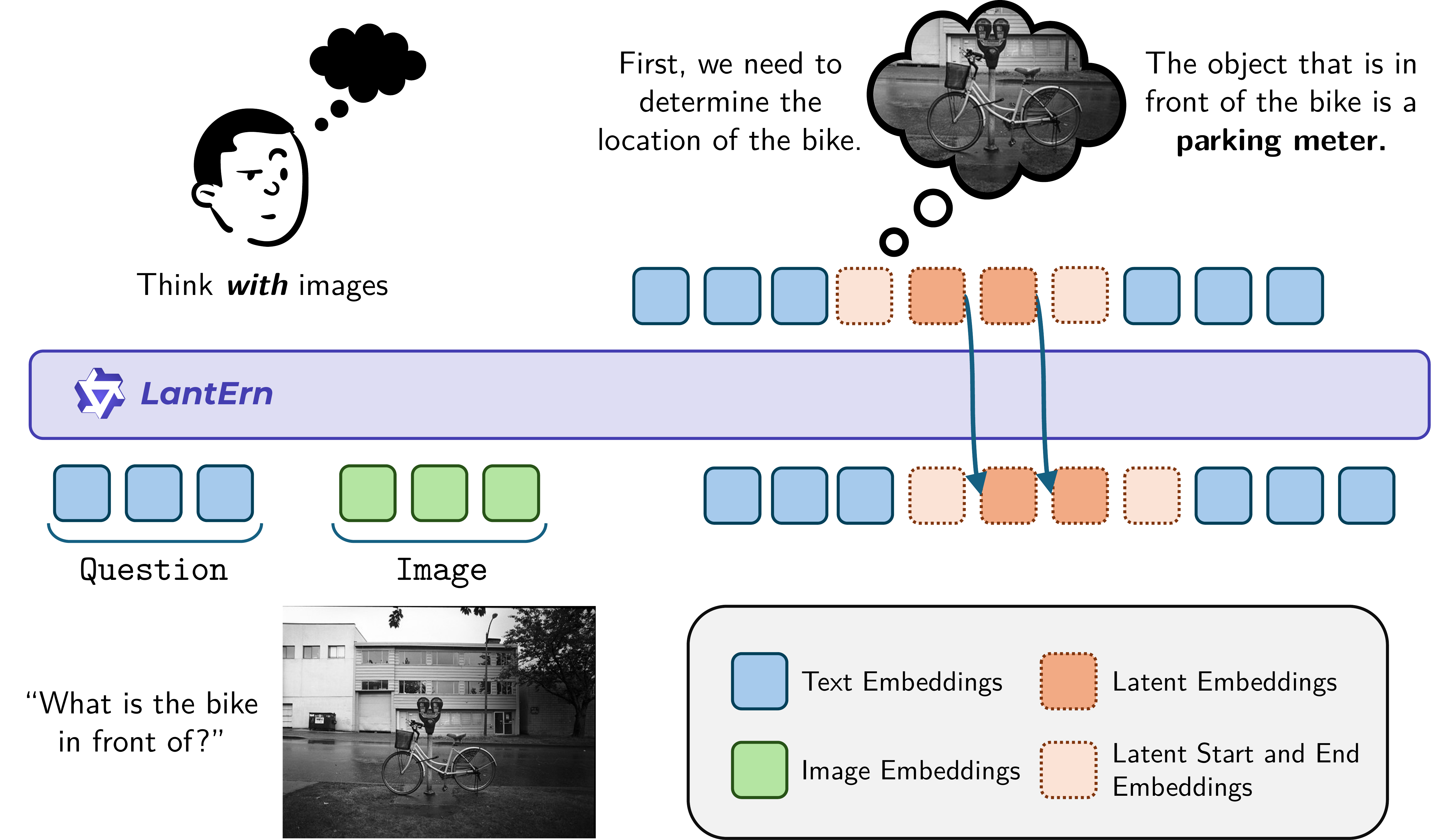}
  \label{fig:lantern-diagram}
  \caption{The \lantern{} framework enables interleaved reasoning between text and latent representations that encode visual ``thoughts''. During inference, \lantern{} can automatically decide when to start latent reasoning by outputting a special token.}
\end{figure}


\section{Introduction}

Large Multimodal Models (LMMs) have achieved strong performance in a wide range of vision-language tasks~\cite{alayrac2022flamingo,liu2023llava,bai2025qwen25vltechnicalreport}, yet their reasoning processes remain predominantly linguistic. In most current systems, visual inputs are encoded once and all subsequent reasoning is carried out in text, a regime referred to as `thinking about images'. 
This forces high-dimensional perceptual information into a low-bandwidth symbolic medium, a limitation that becomes particularly evident on perception-heavy benchmarks, where purely textual chains of thought fail to capture fine-grained spatial and visual structure~\citep{fu2024blink,xiao2024logicvista}.

To overcome these limitations, recent work has shifted toward `thinking \emph{with} images', in which visual information actively participates in the reasoning process rather than being consumed only at the input stage. Existing approaches in this category can be broadly divided into two streams. The first consists of \emph{tool-based visual reasoning} methods, which allow models to invoke external vision modules during inference, such as cropping, object detection, or image generation tools~\citep{yang2023mmreact,suris2023vipergpt,chameleonteam2025chameleonmixedmodalearlyfusionfoundation}. These approaches are limited to a set of predefined tools and often incur significant computational overhead. The second stream performs reasoning by explicitly generating images during the reasoning chain, forcing intermediate visual thoughts to be expressed in pixel space and spending significant computation on photorealistic details that may be irrelevant for the task, which is wasteful ~\citep{chameleonteam2025chameleonmixedmodalearlyfusionfoundation, deng2025emergingpropertiesunifiedmultimodal}.

More recently, \emph{latent visual reasoning} has emerged as an internalized form of thinking with images, in which models maintain and manipulate continuous visual representations in latent space throughout the reasoning process~\citep{li2025lvr,yang2025mirage}. By interleaving latent visual states with text, these methods avoid explicit image generation while preserving visual structure, enabling reasoning to operate over abstract visual representations rather than pixel space.

In this work, we introduce \lantern{} (Latent Visual Structured Reasoning), a framework that enables LMMs to reason using compact latent visual tokens interleaved with language. \lantern{} augments a vision-language transformer with the ability to emit and attend to latent visual states, allowing reasoning to occur directly in the visual feature space of the model. We train \lantern{} in two stages. First, we perform supervised fine-tuning on a custom dataset with annotated visual reasoning traces, grounding latent states in the outputs of the model’s vision encoder. Second, we apply reinforcement learning to optimize both textual and latent reasoning as a sequential decision-making process, using final answer correctness as the reward signal.

We evaluate \lantern{} on challenging visual reasoning benchmarks, including Visual-CoT~\citep{shao2024visualcot}, $V^\star$~\citep{cheng2025vstar}, and Blink~\citep{fu2024blink}. Across all settings, latent visual reasoning achieves comparable or superior performance, suggesting that internal visual representations offer a promising direction for multimodal reasoning.

\section{Related Work}

\paragraph{Text-based and tool-based visual reasoning.}
Early multimodal reasoning approaches extended textual chain-of-thought methods to vision-language tasks, reasoning entirely in text after a single visual encoding pass~\citep{zhang2023multimodalcot,shao2024visualcot}. To address their perceptual limitations, subsequent work introduced tool-augmented reasoning, enabling models to iteratively interact with images through external operations such as cropping, detection, and image generation~\citep{yang2023mmreact,suris2023vipergpt,chameleonteam2025chameleonmixedmodalearlyfusionfoundation}. While effective, these approaches depend on hand-designed tools and do not learn internal visual abstractions.

\paragraph{Latent visual reasoning.}
Latent visual reasoning aims to internalize perceptual processes by allowing models to generate continuous visual representations during inference. \citet{li2025lvr} introduces Latent Visual Reasoning (LVR), a method that conditions the final answer on latent visual tokens. Similarly, \citet{yang2025mirage} introduce machine mental imagery through latent image representations. These works show that preserving visual information in latent space can improve fine-grained reasoning. However, unlike prior approaches that primarily append latent visual representations to support downstream decoding, \lantern{} formulates this paradigm as an \emph{interleaved} reasoning process that alternates between text and latent visual tokens, allowing iterative refinement of internal visual representations and tighter coupling between perception and language.

\section{Our Method: \lantern{}}\label{sec:methodology}

\lantern{} provides a structured mechanism for incorporating visually grounded latent states alongside textual reasoning in LMMs. Standard LMMs are constrained to verbalize every step of visual processing, often forcing high-dimensional visual information into low-bandwidth natural language. We propose a modeling approach where the model learns to generate compressed, non-verbal ``thought'' vectors derived directly from visual features, interleaving these latent states with discrete text generation.

\subsection{Modeling Latent Visual Reasoning}

We model reasoning as a hybrid trajectory $\tau = [s_1, s_2, \dots, s_T]$, where each state $s_t$ lies in either the discrete vocabulary space $\mathcal{V}$ or the continuous latent space $\mathbb{R}^d$, with $d$ denoting the model’s hidden dimension. At each step $t$, the model outputs either a token $w_t \in \mathcal{V}$ (text mode) or a latent vector $\mathbf{z}_t \in \mathbb{R}^d$ (visual latent mode).

To implement this, we build on the Qwen2.5-VL architecture~\citep{bai2025qwen25vltechnicalreport} and extend its vocabulary with three control tokens: \texttt{<|lvr\_start|>}, \texttt{<|lvr\_sep|>}, and \texttt{<|lvr\_end|>}. These tokens act as gating signals that regulate transitions between operating modes:

\begin{enumerate}[leftmargin=*, itemsep=0pt, topsep=5pt,parsep=15pt]
    \item \textbf{Text Mode:} The model functions as a standard autoregressive language model. The hidden state at time $t$, $\mathbf{h}_t$, is passed through the language modeling head to predict a probability distribution over the vocabulary $\mathcal{V}$.
    \item \textbf{Visual Latent Mode:} Upon generating \texttt{<|lvr\_start|>}, for the subsequent $K$ time steps (where $K$ is a fixed latent size hyperparameter), the model bypasses the language modeling head and outputs the unprojected hidden states of the final transformer layer. These $K$ vectors, denoted $\mathbf{z}_{1}, \mathbf{z}_{2}, \dots, \mathbf{z}_{K}$ constitute a block of latent ``thought'' embeddings. After $K$ steps, a terminating token \texttt{<|lvr\_end|>} is introduced and the model returns to text mode. The latent vectors serve as internal reasoning context, allowing the model to attend to its own high-dimensional visual thoughts without having to verbalize that information into text.
\end{enumerate}

\subsection{Supervised Fine-Tuning: Grounding Latent States in Vision}

A fundamental challenge in latent reasoning is defining a ground truth for the model's internal visual thoughts. Since human annotators cannot provide high-dimensional vector supervision, we devise a strategy to \emph{ground} the latent states using the model’s own visual encoder as a teacher signal.

\subsubsection{Visual Feature Extraction as Supervision} 

We leverage the pre-trained vision encoder of the base model as a teacher for the LMM latent states. Consider a training sample $(I, Q, A, \mathcal{T})$ consisting of an image $I$, a question $Q$, the correct answer $A$, and a human-written reasoning trace $\mathcal{T}$ that references certain visual regions. Let $B$ be the set of bounding boxes in $\mathcal{T}$ corresponding to relevant regions of $I$ that the reasoning trace attends to. For each reasoning step associated with a region $b \in B$, we construct a target latent representation $\mathbf{Z}_{\text{target}}$ as follows:
\begin{equation}
    \mathbf{F}_{b} = \text{VisionEncoder}(I, b)\,,
\end{equation}
\begin{equation}
    \mathbf{Z}_{\text{target}} = \text{Pool}(\mathbf{F}_{b}) \;\in\; \mathbb{R}^{K \times d}\,,
\label{eq:latent_tokens}
\end{equation}
where $\mathbf{F}_{b}$ is the feature map extracted from the vision encoder (e.g. patch embeddings) corresponding to region $b$. To produce the latent tokens, we follow the approach of ~\cite{yang2025mirage} and apply an average pooling operation to $\mathbf{F}_{b}$ to produce a fixed-size sequence of $K$ pooled feature vectors. This sequence $\mathbf{Z}_{\text{target}} = [\mathbf{z}_{\text{target}}^{(1)}, \dots, \mathbf{z}_{\text{target}}^{(K)}]$ serves as the target for the model’s latent block, effectively capturing an aggregated representation of the visual region $b$ as perceived by the vision encoder.

\subsubsection{Hybrid Objective Function}

We train \lantern{} with a multi-task objective that jointly optimizes for language generation and latent visual alignment. The total loss is a sum of two components:
\begin{equation}
    \mathcal{L}_{\lantern{}} = \mathcal{L}_{\text{text}} \;+\; \gamma\, \mathcal{L}_{\text{latent}}\,,
\end{equation}
where $\gamma$ is a weighting hyperparameter. 

\paragraph{Text Generation Loss ($\mathcal{L}_{\text{text}}$).} For standard (non-latent) tokens, we use a cross-entropy loss on the next-token prediction, as in conventional language model fine-tuning. This term $\mathcal{L}_{\text{text}}$ preserves the model’s verbal fluency and ensures it can articulate answers correctly in natural language.

\paragraph{Latent Alignment Loss ($\mathcal{L}_{\text{latent}}$).} For each block of $K$ latent tokens generated between a \texttt{<|lvr\_start|>} and \texttt{<|lvr\_end|>} marker, we apply a regression loss that encourages these latent vectors to match the visual encoder’s features for the corresponding region of interest. Let $\mathbf{H}_{\text{gen}} = [\mathbf{h}_{\text{gen}}^{(1)}, \dots, \mathbf{h}_{\text{gen}}^{(K)}]$ be the sequence of $K$ hidden states produced by the LLM in latent mode, and $\mathbf{Z}_{\text{target}} = [\mathbf{z}_{\text{target}}^{(1)}, \dots, \mathbf{z}_{\text{target}}^{(K)}]$ be the pooled target embeddings as defined above. We minimize the mean-squared error (MSE) between these two sequences:
\begin{equation}
    \mathcal{L}_{\text{latent}} \;=\; \frac{1}{K} \sum_{i=1}^{K} \Big\|\,\mathbf{h}_{\text{gen}}^{(i)} - \mathbf{z}_{\text{target}}^{(i)}\Big\|_2^2\,.
\end{equation}

This encourages the LLM to autoregressively predict the latent representations computed in  Eq.~\eqref{eq:latent_tokens}. By learning to minimize the discrepancy $||\mathbf{h}_{\text{gen}} - \mathbf{z}_{\text{target}}||^2$, the model is effectively learning to ``imagine’’ the visual content in its latent space, reconstructing the key visual features needed to answer the question. 
The goal of this phase is to distill latent visual reasoning capabilities into the LLM through explicit supervision of a predefined set of latent visual thoughts. This trains the model to perform interleaved reasoning over text and latent visual representations, grounding its explanations in internal visual states without requiring it to verbalize every aspect of its visual “thoughts.” However, this supervised objective primarily encourages representational fidelity, motivating the subsequent stage to further align free-form latent reasoning with task-level utility.

\subsection{Reinforcement Learning}
While SFT grounds latent representations in visual features, it primarily enforces a \emph{reconstruction} objective. This can lead to latent representations that match perceptual representations, yet suboptimal for downstream reasoning. We therefore explore Reinforcement Learning (RL) as a free-form mechanism to align latent visual reasoning with \emph{task utility}, rather than visual fidelity alone. We formulate latent visual reasoning as a sequential decision-making problem and investigate whether policy optimization can encourage the model to generate latent states that improve final answer correctness, similar in spirit to outcome-driven RL for language models~\citep{ouyang2022training,schulman2017ppo}. 

\paragraph{Policy Optimization with Hybrid Action Spaces.}
We adopt Group Relative Policy Optimization (GRPO)~\citep{deepseek-math} as our primary optimization algorithm. GRPO is a PPO-style method that stabilizes training by normalizing rewards within sampled groups, and has recently been shown to be effective for reasoning-heavy language model fine-tuning~\citep{guo2025deepseekr1nature,olmo2025olmo3}. Unlike standard RL for LLMs, which operates over a discrete token space, interleaved latent reasoning is naturally formulated as a \emph{hybrid action space}: actions correspond either to discrete text tokens or to continuous latent vectors $\mathbf{z} \in \mathbb{R}^d$. This raises a conceptual challenge, as policy gradient objectives are traditionally defined over categorical distributions. Rather than defining an explicit probability density over latent vectors, we treat latent generation as an intermediate computation that conditions subsequent text generation, following prior work on differentiable latent reasoning in language models~\citep{li2025lvr,yang2025mirage}.

Under this formulation, optimization is applied only to the likelihood of discrete text tokens, while gradients propagate through the latent states via standard backpropagation. This design implicitly encourages the model to generate latent representations that improve downstream token predictions and, ultimately, task reward.

\subsubsection{Latent-Aware GRPO Objective}

Given an input query $q$ and image $I$, we sample a group of $G$ rollouts $\{o_1, \dots, o_G\}$ from the old policy $\pi_{\theta_{\text{old}}}$. Each rollout consists of interleaved text and latent blocks. We define the GRPO objective over the discrete text tokens while treating latent states as contextual conditioning variables:
\begin{equation*}
\tilde{L}_{i,t}(\theta)=
\min\!\Big(
r_{i,t}(\theta)\hat{A}_{i},\;
\mathrm{clip}(r_{i,t}(\theta),1-\epsilon,1+\epsilon)\hat{A}_{i}
\Big)
\end{equation*}
\begin{equation}
\mathcal{J}(\theta)=
\mathbb{E}_{\{o_i\}\sim\pi_{\theta_{\mathrm{old}}}}
\!\left[
\frac{1}{G}\sum_{i=1}^{G}\frac{1}{|o_i|}
\sum_{t=1}^{|o_i|}
\bigg(\tilde{L}_{i,t}(\theta)-\beta\,D_{\mathrm{KL}}\big(\pi_\theta(\cdot)\,\|\,\pi_{\mathrm{ref}}(\cdot)\big)\bigg) \,\,\bigg|\,\, q,I
\right],
\end{equation}
where $\pi_\theta(y_{i,t}) \equiv \pi_\theta(y_{i,t}\mid q,I,\tilde h_i^{\mathrm{latent}},y_{i,<t})$, $r_{i,t}(\theta)=\pi_\theta(y_{i,t})/\pi_{\theta_{\mathrm{old}}}(y_{i,t})$ is the importance ratio, and 
$\hat{A}_{i}$ is the group-normalized advantage. 

\paragraph{Latent State Replay.} 
A practical challenge arises from the fact that latent vectors are generated dynamically by the policy. Small parameter updates can lead to significant drift in latent trajectories, which destabilizes the importance sampling ratio. To mitigate this effect, we employ \emph{latent state replay}: during policy updates, the model is forced to condition on the exact latent vectors $\mathbf{H}_{\text{rollout}}$ generated during sampling. This ensures that probability ratios reflect changes in the text policy under a fixed internal reasoning trace, while still allowing gradients to flow back to the parameters responsible for producing latent states.

\subsubsection{Reward Design}

Since there is no direct supervision for the quality of latent thoughts, we rely on sparse outcome-based rewards combined with structural constraints. The total reward is defined as a weighted sum:

\begin{enumerate}[leftmargin=*, itemsep=0pt, topsep=2pt]
    \item \textbf{Accuracy Reward ($R_{\text{acc}}$):} A binary reward indicating whether the final textual answer matches the ground truth. This sparse signal serves as the primary driver of task-oriented latent reasoning, consistent with prior RL-based approaches to reasoning supervision~\citep{ouyang2022training,li2025lvr}.
    \item \textbf{Format Reward ($R_{\text{fmt}}$):} A structural reward that encourages the explicit use of a set of tags, such as \texttt{<think>}, \texttt{</think>} for the reasoning chain, latent reasoning delimiters \texttt{<|lvr\_start|>} and \texttt{<|lvr\_end|>} and \texttt{<|answer|>}, \texttt{</answer>}  for the final answer. This discourages collapse to purely textual reasoning and enforces the presence of a latent block.
\end{enumerate}

Overall, this RL stage is intended to test the hypothesis that outcome-driven optimization can shift latent representations from merely encoding visual appearance toward selectively representing task-critical visual information. Rather than assuming that visually faithful latent states are optimal, we explore whether RL can induce more abstract and utility-driven internal visual reasoning.
\section{Experiments}

We evaluate \lantern{} through a two-stage training pipeline. First, we perform supervised fine-tuning (SFT) to initialize latent visual reasoning by grounding latent states in perceptual features. Second, we apply reinforcement learning (RL) to evaluate whether outcome-driven optimization can further refine the model’s latent reasoning capabilities.

\subsection{Supervised Fine-Tuning Setup}

\paragraph{Dataset Construction.}
\label{dataset}
To train the model to associate latent states with visually relevant regions, we construct a synthetic dataset derived from Visual-CoT~\citep{shao2024visualcot}. Visual-CoT provides image–question pairs accompanied by detailed reasoning traces and a bounding box highlighting the region of the image the model should focus on in order to answer the question effectively, making it a suitable foundation for structured latent supervision.

We employ a large-scale reasoning-oriented multimodal model, \textbf{Qwen3-VL-235B-Thinking} ~\citep{yang2025qwen3technicalreport}, as the reference model. For each image–question pair and its auxiliary images, we prompt the model to generate a structured reasoning trace consisting of three components:
\begin{enumerate}[leftmargin=*, itemsep=0pt, topsep=2pt]
    \item \textbf{Pre-visual thought:} A textual plan describing the visual information required (e.g., \emph{“I need to identify the game title shown on the screen”}).
    \item \textbf{Visual grounding:} A set of bounding boxes highlighting regions of interest (ROIs) that are relevant for answering the question (matching the ground-truth boxes).
    \item \textbf{Post-visual thought:} A textual deduction derived specifically from the visual content within those ROIs.
\end{enumerate}
This procedure yields training samples in which bounding boxes are explicitly aligned with individual reasoning steps in the supervision signal. During SFT, these bounding boxes are used exclusively to extract the corresponding target feature representations from the vision encoder, which are then used as regression targets for the model’s latent blocks. Importantly, the bounding boxes themselves are never exposed to the model during training.

\vspace{-2mm}
\paragraph{Training Configuration.}
We initialize all models from Qwen2.5-VL-3B-Instruct~\citep{bai2025qwen25vltechnicalreport}. To study the effect of latent capacity, we train variants with different latent block sizes, each dubbed LantErn-SFT-$\{K\}$, where $K \in \{4, 8, 16, 32\}$. We use a weighting factor $\gamma=0.1$ for the latent alignment loss. These ablations allow us to examine how the dimensionality of latent tokens affect downstream reasoning behavior.

\paragraph{Baselines.}
\label{sft}
\label{sft}
To isolate the contribution of continuous latent reasoning, we also train a version with only next-token prediction, namely LantErn-NTP. This baseline uses the same backbone architecture, same data, and special control tokens (e.g., \texttt{<|lvr\_start|>}, \texttt{<|lvr\_end|>}), but treats the intermediate reasoning sequence as standard discrete text tokens. In particular, these intermediate reasoning tokens are autoregressively predicted using the standard language modeling head, just like regular text tokens. The language modeling head is never bypassed and no latent regression loss is applied. This comparison controls for additional computation steps and token structure, ensuring that any observed differences can be attributed specifically to the presence of continuous latent representations.
We also include Qwen2.5-VL-3B as the base pretrained model without task-specific fine-tuning.
We do not include direct comparisons with previously proposed latent visual reasoning methods (e.g., LVR and Mirage) because they are trained and evaluated with significantly larger backbone models (typically $\sim$7B parameters), whereas our experiments are conducted with a 3B-scale model. Given the substantial impact of model scale on multimodal reasoning performance, such comparisons would not provide a controlled evaluation of the proposed training approach.

\textbf{Hyperparameters.}
All models are trained using AdamW with a learning rate of $1\times10^{-5}$ and a cosine learning rate schedule, together with a warmup ratio of $0.05$. Additionally, the vision encoder is frozen to both simplify training and improve training stability, allowing the model to focus on learning effective latent visual reasoning on top of fixed visual features.

\subsection{Evaluation Benchmarks}

To evaluate \lantern{}'s performance, we used a subset of Visual-CoT and two vision-centric benchmarks $V^\star$ \cite{wu2023vguidedvisualsearch} and a subset of Blink \cite{fu2024blink}. $V^\star$ assesses a model's ability to perform visual search in real-world scenarios, a fundamental capability of human cognitive reasoning process involving visual information. Blink evaluates core visual perception skills in scenarios where solving the task using text alone (textual priors) is extremely challenging. We used a subset of Blink that focuses on object localization and direct attribution, as it closely aligns with the skills learned in the previous stage \ref{sft}.
Together, these benchmarks provide a basis for evaluating the use of latent visual representations.

\subsection{SFT Results}
\label{sft_results}

\begin{table}[t]
\centering
\scriptsize
\setlength{\tabcolsep}{3pt} 
\begin{tabular}{lccccccc}
\toprule
Model & VisCoT & $V^\star$ & $V_{\text{DA}}^\star$ & $V_{\text{RP}}^\star$ & Blink & Blink$_{\text{OL}}$ & Blink$_{\text{RP}}$ \\
\midrule
Qwen2.5-VL-3B & 0.66 & 0.70 & 0.75 & 0.63 &	0.65 &	0.48 & 0.81 \\
\midrule
LantErn-NTP-4 & 0.80 & 0.72	& 0.71 & 0.72	& 0.60 & 0.45 & 0.72 \\ 
LantErn-SFT-4 & 0.80 & 0.62 & 0.68 & 0.57 & 0.61 & 0.51 & 0.72 \\
LantErn-SFT-8 & 0.81 & 0.65 & 0.71 & 0.60 & 0.60 & 0.52 & 0.68\\
LantErn-SFT-16 & 0.80 & 0.60 & 0.65 & 0.55 & 0.54 & 0.53 & 0.55 \\
LantErn-SFT-32 & 0.79 & 0.72 & 0.72 & 0.71 & 0.58 & 0.49 & 0.66 \\
\bottomrule
\end{tabular}
\caption{SFT evaluation results on $V^\star$, Blink and a subset of VisCoT datasets. DA = Direct Attribution, RP = Relative Position, OL = Object Localization.}
\label{tab:stf_results}
\end{table}


As shown in Table~\ref{tab:stf_results}, all \lantern{} variants improve over the Qwen2.5-VL-3B baseline on Visual-CoT, indicating generalization beyond the supervision data. However, these gains are comparable to the text-only LantErn-NTP baseline (0.80), suggesting that supervised latent grounding alone already yields task-level benefits. SFT mainly improves perception-centric skills: for example, LantErn-SFT-8 increases Blink$_{\text{OL}}$ performance from 0.45 to 0.52, indicating stronger object localization and perceptual grounding. In contrast, performance on relational subsets ($V_{\text{RP}}^\star$ and Blink$_{\text{RP}}$) remains similar or worse than LantErn-NTP, suggesting that latent representations are not yet reliably used for complex reasoning.

Another observation is that performance does not increase monotonically with the latent size $K$. For example, larger latent blocks can lead to degradation on some benchmarks (e.g., Blink$_{\text{RP}}$ drops from 0.72 at $K{=}4$ to 0.66 at $K{=}32$), indicating a trade-off between latent capacity and effective reasoning. This highlights a limitation of fixed-size latent reasoning and suggests that future work could benefit from mechanisms that adapt latent capacity to match the task complexity.

Overall, the SFT results suggest that supervised latent grounding provides perceptual structure but is insufficient on its own to produce consistent task-level improvements, motivating the subsequent reinforcement learning stage.

\subsection{Reinforcement Learning Setup}

Following SFT, we apply RL on the LantErn-SFT-8 model to further refine the latent reasoning policy. At the same time we keep the same baselines as before.

\paragraph{Dataset.}
For the RL stage, we use the VIRL-39k dataset~\citep{virl39k}, which contains a diverse collection of visual reasoning problems without explicit region-level supervision. This setting allows us to evaluate whether RL can guide \lantern{} in the absence of bounding box annotations, relying only on task-level feedback.

\paragraph{Implementation Details.}
We implement the RL training loop using the TRL library~\citep{trl2023}, extending the standard \texttt{GRPOTrainer} to support \emph{latent state replay}. As described in Section~\ref{sec:methodology}, latent replay records the continuous hidden states generated during the rollout phase and reinjects them during the policy update step. This modification stabilizes training by ensuring that importance sampling ratios are computed under a fixed latent trajectory, while still allowing gradients to propagate to the parameters responsible for latent generation.

\paragraph{Hyperparameters.}
We train with a learning rate of $5\times10^{-6}$ and a warmup ratio of $0.03$ and latent size $k=8$, as using 8 latent tokens seems to yield the best overall performance, as indicated in Section~\ref{sft_results}). We set the KL regularization coefficient to $\beta=0.1$ to limit policy drift from the SFT initialization.

During the rollout phase, we sample $G=4$ completions per prompt using temperature $T=0.6$ and top-$p=0.85$ to encourage exploration of diverse latent reasoning trajectories. The reward function combines a sparse accuracy reward (weight 1.0) with a format reward (weight 1.0), which encourages the explicit use of latent reasoning blocks and prevents collapse to purely textual solutions. 

\subsection{RL Results}

\begin{table}[t]
\centering
\scriptsize
\setlength{\tabcolsep}{3pt} 
\begin{tabular}{lccccccc}
\toprule
Model & VisCoT & $V^\star$ & $V_{DA}^\star$ & $V_{RP}^\star$ & Blink & Blink$_{OL}$ & Blink$_{RP}$ \\
\midrule
Qwen2.5-VL-3B & 0.66 & 0.70 & 0.75 & 0.63 & 0.65 &	0.48 & \textbf{0.81} \\
\midrule
NTP-RL & 0.82 & 0.66 & 0.75 & 0.57 & 0.64 & 0.47 & 0.80 \\ 
LantErn-RL-8 & \textbf{0.83} & \textbf{0.71} & \textbf{0.76} & \textbf{0.67} & \textbf{0.68} & \textbf{0.54} & \textbf{0.81}  \\
\bottomrule
\end{tabular}
\caption{
RL evaluation results on VStar, Blink and a subset of Viscot datasets. DA = Direct Attribution, RP = Relative Position, OL = Object Localization.}
\label{tab:vstar_blink_detailed}
\end{table}

Applying RL on top of LantErn-SFT-8 leads to consistent performance improvements, outperforming both the base model and the NTP-RL variant, across all evaluated benchmarks. The largest gains appear on \textbf{out-of-distribution, perception-heavy} benchmarks. Notably, performance on Blink$_{\text{RP}}$ improves from 0.68 (SFT) to 0.81, representing a substantial gain.
This trend is consistent across additional tasks: compared to the NTP baseline, performance increases on $V^\star_{\text{RP}}$ (0.57 $\rightarrow$ 0.67) and Blink$_{\text{OL}}$ (0.47 $\rightarrow$ 0.54), further indicating improved spatial and relational reasoning.

These results support our hypothesis that RL is the stage at which latent states transition from perceptually faithful reconstruction to task-driven internal visual representations. Although additional ablations are needed to fully characterize this effect, the consistent gains across benchmarks indicate that RL enables more effective internal use of visual information. Finally, achieving parity with a 7B model on several benchmarks highlights the potential of latent visual reasoning as a compute-efficient alternative to model scaling for perception-centric tasks.

\section{Conclusions}

In this paper, we present \lantern{} as a novel multimodal hybrid reasoning framework that interleaves latent visual reasoning with standard text generation. Our framework is trained in two stages. First, we perform SFT to distill this capability into the model by explicitly joint supervision in the latent representations and text tokens, enabling it to align these latent representations with visual concepts and to form abstract visual thoughts. In the second stage, we apply RL to further train the model to generate its own latent representations without being constrained to maintain strict fidelity enforced by the previous stage. This grants the model greater freedom to explore task-specific solutions and to revamp abstract visual thoughts, resulting in improved performance on visual reasoning benchmarks.

\textbf{Limitations and Future Work}: Despite its effectiveness, the framework has some limitations. First, interleaved latent reasoning depends on the quality and diversity of multimodal trajectories, which are currently concentrated in a narrow visual domain. Second, the model uses a fixed number of latent tokens; enabling dynamically sized latent blocks that adapt to task complexity is a promising direction. Finally, a deeper analysis of latent dependencies is needed, including methods to visualize latent representations and better understand their utility and alignment with generated text.

\section*{Acknowledgments}

This work was supported by the Portuguese Recovery and Resilience Plan through project C645008882-00000055 (Center for ResponsibleAI), by the project DECOLLAGE (ERC-2022-CoG 101088763), and by FCT/MECI through national funds and when applicable co-funded EU funds under UID/50008: Instituto de Telecomunicações.

\section{Ethics Statement}

\lantern{} is a multimodal reasoning framework designed to improve visual reasoning capabilities in large vision–language models. While such systems can enable beneficial applications in accessibility, education, and scientific analysis, they also raise ethical considerations related to misuse, bias, and transparency. Multimodal models may inherit biases present in their training data, including cultural, demographic, or representational imbalances. These biases can affect model outputs and may disproportionately impact underrepresented groups. Although Lantern focuses on reasoning mechanisms rather than dataset expansion, it relies on existing multimodal corpora whose limitations may propagate into the model. Care should be taken when deploying such systems in high-stakes settings.

\section{Reproducibility Statement}

We prioritize reproducibility by providing detailed descriptions of the Lantern architecture, training pipeline, and evaluation protocols. The paper specifies the model backbone, latent reasoning mechanism, and the two-stage training procedure (supervised fine-tuning and reinforcement learning). Hyperparameters, optimization settings, and dataset compositions are reported in the main text and appendix.
We use publicly available benchmarks for evaluation and clearly describe preprocessing and evaluation procedures. All experiments are conducted using deterministic training configurations where possible, including fixed random seeds and documented hardware setups.
To facilitate replication, we plan to release implementation details, training scripts, and configuration files upon publication. These materials will include instructions for reproducing the main experiments, along with pretrained checkpoints where licensing permits.
We acknowledge that training large multimodal models requires significant computational resources. To mitigate this barrier, we provide ablation studies and smaller-scale configurations that reproduce key findings using reduced compute budgets.

\bibliography{iclr2026_conference}

@article{fu2024blink,
          title={BLINK: Multimodal Large Language Models Can See but Not Perceive},
          author={Fu, Xingyu and Hu, Yushi and Li, Bangzheng and Feng, Yu and Wang, Haoyu and Lin, Xudong and Roth, Dan and Smith, Noah A and Ma, Wei-Chiu and Krishna, Ranjay},
          journal={arXiv preprint arXiv:2404.12390},
          year={2024}
        }

@misc{zhang2023multimodalcot,
      title={Multimodal Chain-of-Thought Reasoning in Language Models}, 
      author={Zhuosheng Zhang and Aston Zhang and Mu Li and Hai Zhao and George Karypis and Alex Smola},
      year={2024},
      eprint={2302.00923},
      archivePrefix={arXiv},
      primaryClass={cs.CL},
      url={https://arxiv.org/abs/2302.00923}, 
}

@inproceedings{
shao2024visualcot,
title={Visual CoT: Advancing Multi-Modal Language Models with a Comprehensive Dataset and Benchmark for Chain-of-Thought Reasoning},
author={Hao Shao and Shengju Qian and Han Xiao and Guanglu Song and Zhuofan Zong and Letian Wang and Yu Liu and Hongsheng Li},
booktitle={The Thirty-eight Conference on Neural Information Processing Systems Datasets and Benchmarks Track},
year={2024},
url={https://openreview.net/forum?id=aXeiCbMFFJ}
}

@misc{yang2023mmreact,
      title={MM-REACT: Prompting ChatGPT for Multimodal Reasoning and Action}, 
      author={Zhengyuan Yang and Linjie Li and Jianfeng Wang and Kevin Lin and Ehsan Azarnasab and Faisal Ahmed and Zicheng Liu and Ce Liu and Michael Zeng and Lijuan Wang},
      year={2023},
      eprint={2303.11381},
      archivePrefix={arXiv},
      primaryClass={cs.CV},
      url={https://arxiv.org/abs/2303.11381}, 
}

@INPROCEEDINGS{suris2023vipergpt,
  author={Surís, Dídac and Menon, Sachit and Vondrick, Carl},
  booktitle={2023 IEEE/CVF International Conference on Computer Vision (ICCV)}, 
  title={ViperGPT: Visual Inference via Python Execution for Reasoning}, 
  year={2023},
  volume={},
  number={},
  pages={11854-11864},
  doi={10.1109/ICCV51070.2023.01092}}

@misc{chameleonteam2025chameleonmixedmodalearlyfusionfoundation,
      title={Chameleon: Mixed-Modal Early-Fusion Foundation Models}, 
      author={{Chameleon Team}},
      year={2025},
      eprint={2405.09818},
      archivePrefix={arXiv},
      primaryClass={cs.CL},
      url={https://arxiv.org/abs/2405.09818}, 
}

@misc{li2025lvr,
      title={Latent Visual Reasoning}, 
      author={Bangzheng Li and Ximeng Sun and Jiang Liu and Ze Wang and Jialian Wu and Xiaodong Yu and Hao Chen and Emad Barsoum and Muhao Chen and Zicheng Liu},
      year={2025},
      eprint={2509.24251},
      archivePrefix={arXiv},
      primaryClass={cs.CV},
      url={https://arxiv.org/abs/2509.24251}, 
}

@misc{yang2025mirage,
      title={Machine Mental Imagery: Empower Multimodal Reasoning with Latent Visual Tokens}, 
      author={Zeyuan Yang and Xueyang Yu and Delin Chen and Maohao Shen and Chuang Gan},
      year={2025},
      eprint={2506.17218},
      archivePrefix={arXiv},
      primaryClass={cs.CV},
      url={https://arxiv.org/abs/2506.17218}, 
}

@misc{cheng2025vstar,
      title={V-STaR: Benchmarking Video-LLMs on Video Spatio-Temporal Reasoning}, 
      author={Zixu Cheng and Jian Hu and Ziquan Liu and Chenyang Si and Wei Li and Shaogang Gong},
      year={2025},
      eprint={2503.11495},
      archivePrefix={arXiv},
      primaryClass={cs.CV},
      url={https://arxiv.org/abs/2503.11495}, 
}

@misc{deepseek-math,
      title={DeepSeekMath: Pushing the Limits of Mathematical Reasoning in Open Language Models}, 
      author={Zhihong Shao and Peiyi Wang and Qihao Zhu and Runxin Xu and Junxiao Song and Xiao Bi and Haowei Zhang and Mingchuan Zhang and Y. K. Li and Y. Wu and Daya Guo},
      year={2024},
      eprint={2402.03300},
      archivePrefix={arXiv},
      primaryClass={cs.CL},
      url={https://arxiv.org/abs/2402.03300}, 
}

@misc{bai2025qwen25vltechnicalreport,
      title={Qwen2.5-VL Technical Report}, 
      author={Shuai Bai and Keqin Chen and Xuejing Liu and Jialin Wang and Wenbin Ge and Sibo Song and Kai Dang and Peng Wang and Shijie Wang and Jun Tang and Humen Zhong and Yuanzhi Zhu and Mingkun Yang and Zhaohai Li and Jianqiang Wan and Pengfei Wang and Wei Ding and Zheren Fu and Yiheng Xu and Jiabo Ye and Xi Zhang and Tianbao Xie and Zesen Cheng and Hang Zhang and Zhibo Yang and Haiyang Xu and Junyang Lin},
      year={2025},
      eprint={2502.13923},
      archivePrefix={arXiv},
      primaryClass={cs.CV},
      url={https://arxiv.org/abs/2502.13923}, 
}

@inproceedings{ouyang2022training,
 author = {Ouyang, Long and Wu, Jeffrey and Jiang, Xu and Almeida, Diogo and Wainwright, Carroll and Mishkin, Pamela and Zhang, Chong and Agarwal, Sandhini and Slama, Katarina and Ray, Alex and Schulman, John and Hilton, Jacob and Kelton, Fraser and Miller, Luke and Simens, Maddie and Askell, Amanda and Welinder, Peter and Christiano, Paul F and Leike, Jan and Lowe, Ryan},
 booktitle = {Advances in Neural Information Processing Systems},
 editor = {S. Koyejo and S. Mohamed and A. Agarwal and D. Belgrave and K. Cho and A. Oh},
 pages = {27730--27744},
 publisher = {Curran Associates, Inc.},
 title = {Training language models to follow instructions with human feedback},
 url = {https://proceedings.neurips.cc/paper_files/paper/2022/file/b1efde53be364a73914f58805a001731-Paper-Conference.pdf},
 volume = {35},
 year = {2022}
}

@misc{schulman2017ppo,
      title={Proximal Policy Optimization Algorithms}, 
      author={John Schulman and Filip Wolski and Prafulla Dhariwal and Alec Radford and Oleg Klimov},
      year={2017},
      eprint={1707.06347},
      archivePrefix={arXiv},
      primaryClass={cs.LG},
      url={https://arxiv.org/abs/1707.06347}, 
}

@misc{trl2023,
  title={TRL: Transformer Reinforcement Learning},
  author={von Werra, Leandro and Belkada, Younes and Sanh, Victor and others},
  year={2023},
  howpublished={\url{https://github.com/huggingface/trl}}
}

@article{virl39k,
      title={VL-Rethinker: Incentivizing Self-Reflection of Vision-Language Models with Reinforcement Learning},
      author = {Wang, Haozhe and Qu, Chao and Huang, Zuming and Chu, Wei and Lin,Fangzhen and Chen, Wenhu},
      journal={arXiv preprint arXiv:2504.08837},
      year={2025}
}

@misc{yang2025qwen3technicalreport,
      title={Qwen3 Technical Report}, 
      author={An Yang and Anfeng Li and Baosong Yang and Beichen Zhang and Binyuan Hui and Bo Zheng and Bowen Yu and Chang Gao and Chengen Huang and Chenxu Lv and Chujie Zheng and Dayiheng Liu and Fan Zhou and Fei Huang and Feng Hu and Hao Ge and Haoran Wei and Huan Lin and Jialong Tang and Jian Yang and Jianhong Tu and Jianwei Zhang and Jianxin Yang and Jiaxi Yang and Jing Zhou and Jingren Zhou and Junyang Lin and Kai Dang and Keqin Bao and Kexin Yang and Le Yu and Lianghao Deng and Mei Li and Mingfeng Xue and Mingze Li and Pei Zhang and Peng Wang and Qin Zhu and Rui Men and Ruize Gao and Shixuan Liu and Shuang Luo and Tianhao Li and Tianyi Tang and Wenbiao Yin and Xingzhang Ren and Xinyu Wang and Xinyu Zhang and Xuancheng Ren and Yang Fan and Yang Su and Yichang Zhang and Yinger Zhang and Yu Wan and Yuqiong Liu and Zekun Wang and Zeyu Cui and Zhenru Zhang and Zhipeng Zhou and Zihan Qiu},
      year={2025},
      eprint={2505.09388},
      archivePrefix={arXiv},
      primaryClass={cs.CL},
      url={https://arxiv.org/abs/2505.09388}, 
}

@inproceedings{alayrac2022flamingo,
 author = {Alayrac, Jean-Baptiste and Donahue, Jeff and Luc, Pauline and Miech, Antoine and Barr, Iain and Hasson, Yana and Lenc, Karel and Mensch, Arthur and Millican, Katherine and Reynolds, Malcolm and Ring, Roman and Rutherford, Eliza and Cabi, Serkan and Han, Tengda and Gong, Zhitao and Samangooei, Sina and Monteiro, Marianne and Menick, Jacob L and Borgeaud, Sebastian and Brock, Andy and Nematzadeh, Aida and Sharifzadeh, Sahand and Bi\'{n}kowski, Miko\l aj and Barreira, Ricardo and Vinyals, Oriol and Zisserman, Andrew and Simonyan, Kar\'{e}n},
 booktitle = {Advances in Neural Information Processing Systems},
 editor = {S. Koyejo and S. Mohamed and A. Agarwal and D. Belgrave and K. Cho and A. Oh},
 pages = {23716--23736},
 publisher = {Curran Associates, Inc.},
 title = {Flamingo: a Visual Language Model for Few-Shot Learning},
 volume = {35},
 year = {2022}
}

@inproceedings{liu2023llava,
  title     = {Visual Instruction Tuning},
  author    = {Liu, Haotian and Li, Chunyuan and Wu, Qingyang and Lee, Yong Jae},
  booktitle = {Advances in Neural Information Processing Systems},
  year      = {2023},
  url       = {https://arxiv.org/abs/2304.08485}
}

@misc{xiao2024logicvista,
  title         = {LogicVista: Multimodal LLM Logical Reasoning Benchmark in Visual Contexts},
  author        = {Yijia Xiao and Edward Sun and Tianyu Liu and Wei Wang},
  year          = {2024},
  eprint        = {2407.04973},
  archivePrefix = {arXiv},
  primaryClass  = {cs.AI},
  doi           = {10.48550/arXiv.2407.04973},
  url           = {https://arxiv.org/abs/2407.04973}
}

@article{guo2025deepseekr1nature,
  title={DeepSeek-R1 incentivizes reasoning in LLMs through reinforcement learning},
  author={Guo, Daya and Yang, Dejian and Zhang, Haowei and Song, Junxiao and Wang, Peiyi and Zhu, Qihao and Xu, Runxin and Zhang, Ruoyu and Ma, Shirong and Bi, Xiao and others},
  journal={Nature},
  volume={645},
  number={8081},
  pages={633--638},
  year={2025},
  publisher={Nature Publishing Group UK London}
}

@misc{olmo2025olmo3,
      title={Olmo 3}, 
      author={Team Olmo and : and Allyson Ettinger and Amanda Bertsch and Bailey Kuehl and David Graham and David Heineman and Dirk Groeneveld and Faeze Brahman and Finbarr Timbers and Hamish Ivison and Jacob Morrison and Jake Poznanski and Kyle Lo and Luca Soldaini and Matt Jordan and Mayee Chen and Michael Noukhovitch and Nathan Lambert and Pete Walsh and Pradeep Dasigi and Robert Berry and Saumya Malik and Saurabh Shah and Scott Geng and Shane Arora and Shashank Gupta and Taira Anderson and Teng Xiao and Tyler Murray and Tyler Romero and Victoria Graf and Akari Asai and Akshita Bhagia and Alexander Wettig and Alisa Liu and Aman Rangapur and Chloe Anastasiades and Costa Huang and Dustin Schwenk and Harsh Trivedi and Ian Magnusson and Jaron Lochner and Jiacheng Liu and Lester James V. Miranda and Maarten Sap and Malia Morgan and Michael Schmitz and Michal Guerquin and Michael Wilson and Regan Huff and Ronan Le Bras and Rui Xin and Rulin Shao and Sam Skjonsberg and Shannon Zejiang Shen and Shuyue Stella Li and Tucker Wilde and Valentina Pyatkin and Will Merrill and Yapei Chang and Yuling Gu and Zhiyuan Zeng and Ashish Sabharwal and Luke Zettlemoyer and Pang Wei Koh and Ali Farhadi and Noah A. Smith and Hannaneh Hajishirzi},
      year={2025},
      eprint={2512.13961},
      archivePrefix={arXiv},
      primaryClass={cs.CL},
      url={https://arxiv.org/abs/2512.13961}, 
}

@misc{wu2023vguidedvisualsearch,
      title={V*: Guided Visual Search as a Core Mechanism in Multimodal LLMs}, 
      author={Penghao Wu and Saining Xie},
      year={2023},
      eprint={2312.14135},
      archivePrefix={arXiv},
      primaryClass={cs.CV},
      url={https://arxiv.org/abs/2312.14135}, 
}

@misc{deng2025emergingpropertiesunifiedmultimodal,
      title={Emerging Properties in Unified Multimodal Pretraining}, 
      author={Chaorui Deng and Deyao Zhu and Kunchang Li and Chenhui Gou and Feng Li and Zeyu Wang and Shu Zhong and Weihao Yu and Xiaonan Nie and Ziang Song and Guang Shi and Haoqi Fan},
      year={2025},
      eprint={2505.14683},
      archivePrefix={arXiv},
      primaryClass={cs.CV},
      url={https://arxiv.org/abs/2505.14683}, 
}
\bibliographystyle{iclr2026_conference}


\end{document}